\documentclass{article}

\usepackage[final]{nips_2016}

\usepackage[utf8]{inputenc} 
\usepackage[T1]{fontenc}    
\usepackage{url}            
\usepackage{booktabs}       
\usepackage{amsfonts}       
\usepackage{nicefrac}       
\usepackage{microtype}      
\usepackage{graphicx}

\title{Embedding Projector: Interactive Visualization and Interpretation of Embeddings}

\author{
  Daniel Smilkov\\
  Google Brain\\
  \texttt{smilkov@google.com}
  \And
  Nikhil Thorat \\
  Google Brain\\
  \texttt{nsthorat@google.com}
  \And
  Charles Nicholson \\
  Google Brain\\
  \texttt{nicholsonc@google.com}
  \AND
  Emily Reif \\
  Brown University \\
  \texttt{ereif@cs.brown.edu}
  \And
  Fernanda B. Vi\'{e}gas \\
  Google Brain\\
  \texttt{viegas@google.com}
  \And
  Martin Wattenberg \\
  Google Brain\\
  \texttt{wattenberg@google.com}
}

\begin{document}

\maketitle

\begin{abstract}
Embeddings are ubiquitous in machine learning, appearing in recommender systems, NLP, and many other applications. Researchers and developers often need to explore the properties of a specific embedding, and one way to analyze embeddings is to visualize them. We present the Embedding Projector, a tool for interactive visualization and interpretation of embeddings.
\end{abstract}

\section{Introduction}
An embedding is a map from input data to points in Euclidean space. 
Machine learning researchers and developers often need to explore the properties of a specific embedding to understand the behavior of their model. An engineer who creates an embedding of songs for a recommendation system might want to verify that the nearest neighbors of "Stairway to Heaven" include "Whole Lotta Love" and not "Let It Go" from Frozen. Meanwhile, a researcher may be interested in global geometric properties, such as linear relationships between meaningful subsets of embedded points. For both sets of users, gaining an understanding of embedding geometry is a key step in interpreting a machine learning model.

An appealing approach to analyzing embeddings is to visualize them. Since embeddings often exist in a space of hundreds of dimensions, an essential step is "dimensionality reduction" which projects points to a more approachable two or three dimensions. Many tools exist to perform various types of dimensionality reduction, but they are largely non-interactive (e.g., Matplotlib [3], or the code released in [5]).

Unfortunately, these static views are generally inadequate for exploring high-dimensional data--although an important first step in the process, users then typically want to switch quickly between many views, zooming and filtering, and then closely inspecting details. Indeed, there have been many toolkits created with the express purpose of exploring high-dimensional data (e.g., [2], [8]), and a key lesson has been that rich interactivity coupled with multiple linked views is extremely helpful.

At the same time, the embeddings that arise in machine learning differ from the kind of data sets seen in conventional "high-dimensional" visualization. Most notably, traditional systems often assume the underlying dimensions have a particular meaning (the expression level of a given gene, or the age of a person). On the other hand, the basis vectors in a typical embedding are typically not meaningful--in fact, discovering semantically significant directions can be a goal in itself.

In this paper, we present the Embedding Projector, a system for interactive visualization and analysis of high-dimensional data. Although designed to be a general-purpose tool, it is optimized for the use cases that arise in machine learning and includes special features to explore meaningful directions in a data set. 

\section{How Users (Want to) Interpret Embeddings}

We held informal interviews with engineers and researchers within our organization to learn how they worked (or wanted to work) with embeddings. We also looked at usage of existing internal tools. Three themes arose consistently.

\paragraph{Task 1. Exploring local neighborhoods} Many users wanted to inspect the nearest neighbors of a given point. Confirming that nearby points are semantically related represented an important step in establishing trust in an algorithm.

\paragraph{Task 2. Viewing global geometry and finding clusters} Several users were interested in finding large clusters of related points, as well as seeing the global geometry of the embeddings. 

\paragraph{Task 3. Finding meaningful "directions"} Researchers have found that embedding spaces sometimes contain semantically significant directions. For example [6] discovered that the vector defined by subtracting the point for "man" from the point for "woman" represented a kind of "female" direction in space. Users we talked with considered the discovery of such directions interesting and worthwhile. We know of no current tools that help with this use case.

\section{The Embedding Projector Application}

The Embedding Projector is a web application, available as both a standalone tool and integrated into the TensorFlow platform [1]. Users may either upload arbitrary high-dimensional data, in a simple text format, or (in TensorFlow) take advantage of the model checkpoint system that makes it easy to visualize any tensors as an embedding.

\begin{figure}[ht]
  \centering
  \centerline{\includegraphics[width=\textwidth]{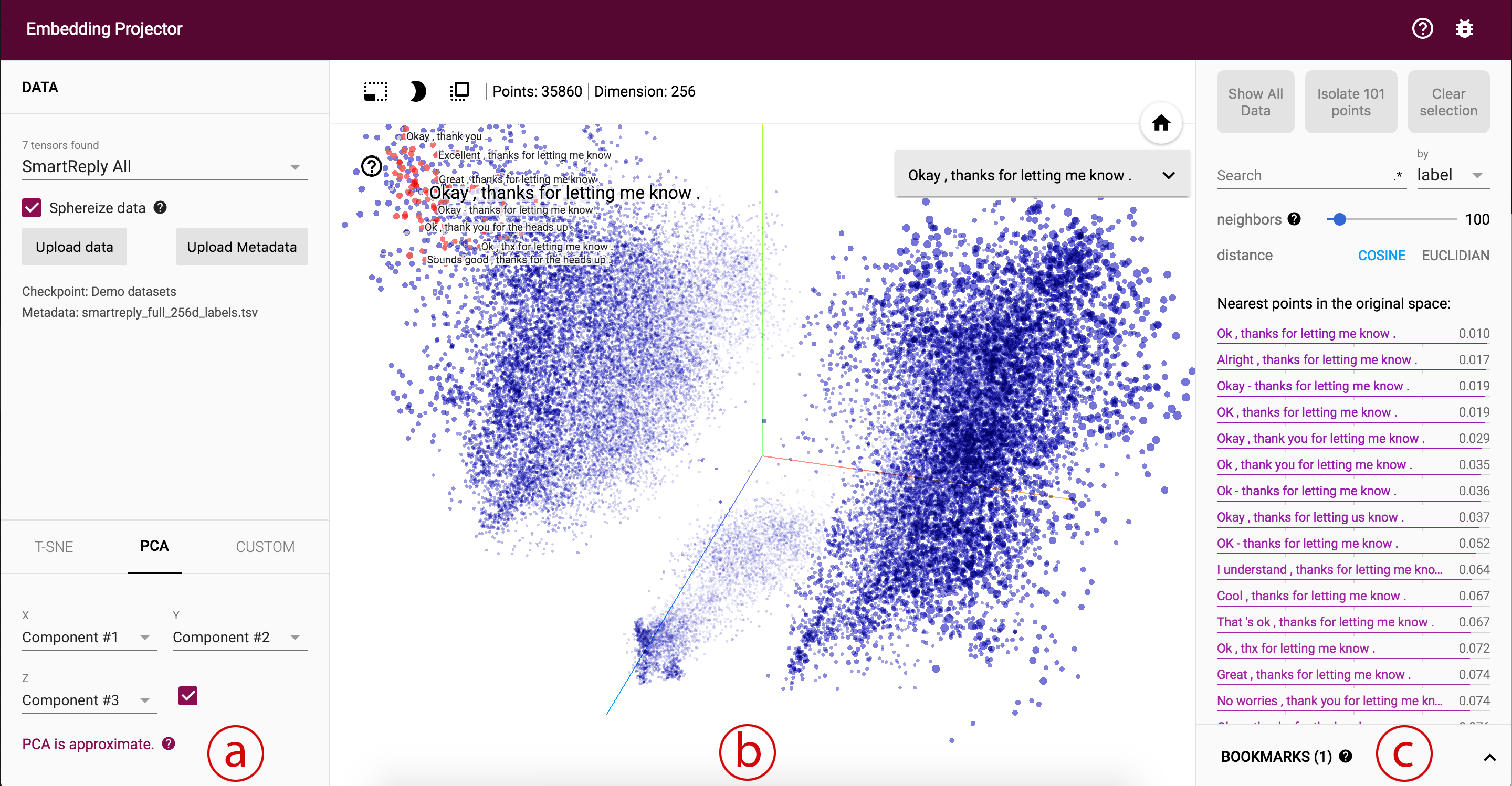}}
  \caption{A PCA projection of a corpus of 35k frequently used phrases in emails [4].}
  \label{fig-main}
\end{figure}

Fig.~\ref{fig-main} shows the main view of the web app with (a) \textit{data} panel on the left, where users can choose data columns to color and label points, (b) the \textit{projected} view in the center, and (c) the \textit{inspector} panel on the right side, where users can search for particular points and see a list of nearest neighbors.



The Embedding Projector offers three methods of reducing the dimensionality of a data set: two linear and one nonlinear. Each method can be used to create either a two- or three-dimensional view.

\paragraph{Principal Component Analysis}
A straightforward technique for reducing dimensions is Principal Component Analysis (PCA). The Embedding Projector computes the top 10 principal components. Menus let the user project those components onto any combination of two or three. PCA is often effective at supporting Task 2, examining global geometry.

\paragraph{t-SNE}
A popular non-linear dimensionality reduction technique is t-SNE [5]. The Embedding Projector offers both two- and three-dimensional t-SNE views. Layout is performed client-side. Because t-SNE often preserves some local structure, in practice it supports both Task 1 and Task 2.

\paragraph{Custom}
Users can construct specialized linear projections based on text searches, supporting Task 3, finding meaningful directions in space. To define a projection axis, the user enters two search strings or regular expressions. The program computes the centroids of the sets of points whose labels match these searches, and uses the difference vector between centroids as a projection axis. 
For example, in the Smart Reply data [4], this view uncovered a surprisingly regular relationship between phrases ending in periods versus exclamation points.


\subsection{Interacting with the visualizations}

To explore a data set, users can navigate the views in either a 2D or a 3D mode, zooming, rotating, and panning using natural click-and-drag gestures. To help interpretation of the 3D mode, the Projector uses multiple redundant depth cues: changing the size of points based on distance to camera; adding fog to fade out more distant points; and initializing the view with an animated "lazy susan" motion. Both modes exploit WebGL to provide smooth, fluid interaction, which encourages exploration of the underlying space, supporting both Task 1 and Task 2. 

To support Task 1, clicking on a point causes the right pane to show an explicit textual list of nearest neighbors, along with distances to the current point. The nearest-neighbor points themselves are highlighted on the projection. 

Users sometimes wish to focus on a subset of points--an interesting cluster, or perhaps the set of nearest neighbors of a given point. Zooming into the cluster gives some information, but it is sometimes more helpful to restrict the view to a subset of points and perform t-SNE or PCA only on those points. To do so, the user can select points in multiple ways: after clicking on a point, its nearest neighbors are selected; after a search, the points matching the query are selected; right-clicking and dragging defines a selection sphere. After selecting a set of points, the user can isolate those points for further analysis on their own with the "Isolate Points" button in the Inspector pane on the right hand side.

\begin{figure}[!tbp]
  \centering
  \begin{minipage}[b]{0.48\textwidth}
    \includegraphics[width=\textwidth]{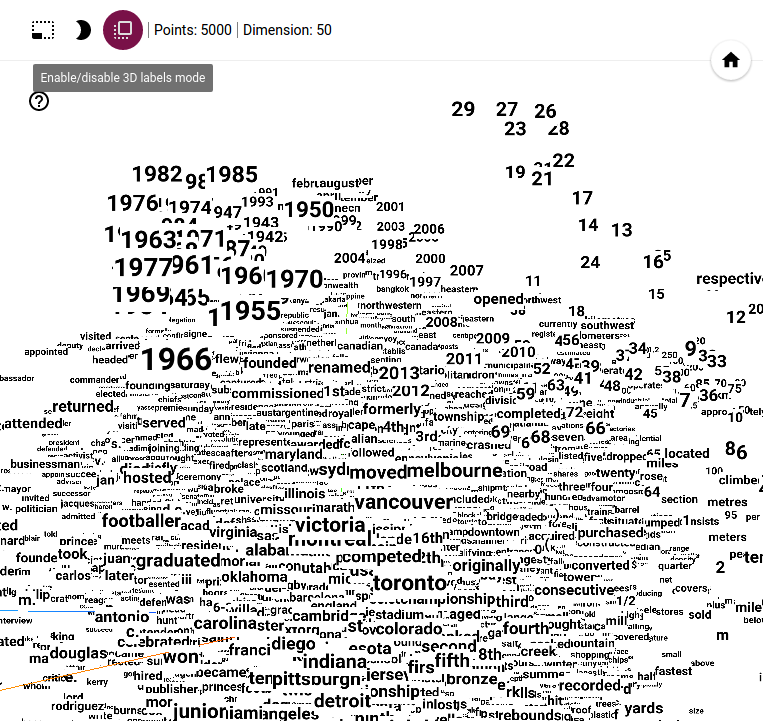}
    \caption{3D labels view of word embeddings.}
    \label{3d-labels}
  \end{minipage}
  \hfill
  \begin{minipage}[b]{0.48\textwidth}
    \includegraphics[width=\textwidth]{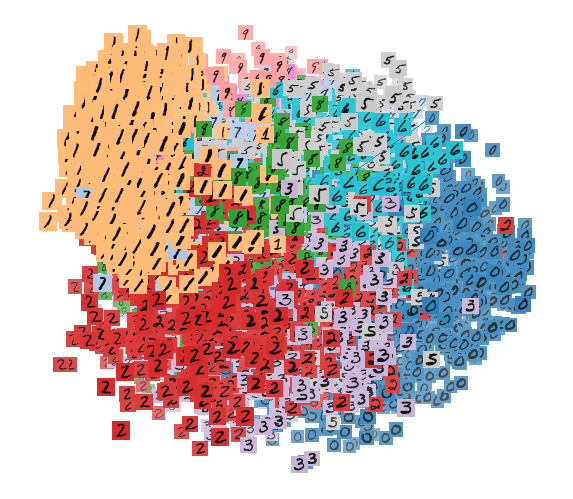}
    \caption{Image view of the MNIST dataset.}
    \label{mnist-colorized}
  \end{minipage}
\end{figure}

In 3D mode, labels can appear as "billboards" -- flat images that always face the user. Fig.~\ref{3d-labels} depicts this view. When there is an image associated with a data point, we can simply use the actual image as the "label". A nice example is an embedding of MNIST images shown in Fig.~\ref{mnist-colorized} where each image has a background color according to its true label.


\paragraph{Collaborative Features}
Early in the development process we saw that users of the Embedding Projector
wanted to share specific views of their data. 
Indeed, previous work in  visualization has shown the benefits of allowing users to share the state of a visualization [7].

To allow easy collaboration, the Embedding Projector lets users save the current state (including computed coordinates of t-SNE embeddings) as a small file. The Projector can then be pointed to a set of one or more of these files, producing the panel seen in the bottom right of Fig.~\ref{fig-main}. Other users can then walk through a sequence of bookmarks.

\section{Conclusion}

The Embedding Projector is a new visualization tool that helps users interpret machine learning models that rely on embeddings. Unlike other high-dimensional visualization systems, it is customized for the kinds of tasks faced by machine learning developers and researchers: exploring local neighborhoods for individual points, analyzing global geometry, and investigating semantically meaningful vectors in embedding space. The Projector is part of the TensorFlow platform, and seamlessly allows analysis and interpretation of TensorFlow models.


There are a number of directions for future work on the visualization. For example, when developing multiple versions of a model, or inspecting how a model changes over time, it could be useful to visually compare two embeddings. Doing so would require nontrivial additions to the current visualizations. A second direction for future research is to make it easier for users to discover meaningful directions in the data. While the current interface makes it easy to explore various hypotheses, there may be ways for the computer to generate and test hypotheses automatically.

\subsubsection*{Acknowledgments}
Thanks to Shan Carter for invaluable design contributions, and to our colleagues who provided thoughtful feedback.

\section*{References}

\small

[1] M.~Abadi, A.~Agarwal, P.~Barham, E.~Brevdo, Z.~Chen, C.~Citro, G.~S. Corrado,
  A.~Davis, J.~Dean, M.~Devin, et~al.
\newblock Tensorflow: Large-scale machine learning on heterogeneous distributed
  systems.
\newblock {\em arXiv preprint arXiv:1603.04467}, 2016.

[2] C.~Ahlberg and B.~Shneiderman.
\newblock Visual information seeking: Tight coupling of dynamic query filters
  with starfield displays.
\newblock In {\em Proceedings of the SIGCHI conference on Human factors in
  computing systems}, pages 313--317. ACM, 1994.

[3] J.~D. Hunter et~al.
\newblock Matplotlib: A 2d graphics environment.
\newblock {\em Computing in science and engineering}, 9(3):90--95, 2007.

[4] A.~Kannan, K.~Kurach, S.~Ravi, T.~Kaufmann, A.~Tomkins, B.~Miklos, G.~Corrado,
  L.~Luk{\'a}cs, M.~Ganea, P.~Young, et~al.
\newblock Smart reply: Automated response suggestion for email.

[5] L.~v.~d. Maaten and G.~Hinton.
\newblock Visualizing data using t-sne.
\newblock {\em Journal of Machine Learning Research}, 9(Nov):2579--2605, 2008.

[6] T.~Mikolov, K.~Chen, G.~Corrado, and J.~Dean.
\newblock Efficient estimation of word representations in vector space.
\newblock {\em arXiv preprint arXiv:1301.3781}, 2013.

[7] F.~B. Viegas, M.~Wattenberg, F.~Van~Ham, J.~Kriss, and M.~McKeon.
\newblock Manyeyes: a site for visualization at internet scale.
\newblock {\em Visualization and Computer Graphics, IEEE Transactions on},
  13(6):1121--1128, 2007.

[8] C.~Weaver.
\newblock Building highly-coordinated visualizations in improvise.
\newblock In {\em Information Visualization, 2004. INFOVIS 2004. IEEE Symposium
  on}, pages 159--166. IEEE, 2004.

\end{document}